\documentclass[final]{cvpr}

\usepackage{times}
\usepackage{epsfig}
\usepackage{graphicx}
\usepackage[dvipsnames]{xcolor}
\usepackage{amsmath}
\usepackage{amssymb}
\usepackage{placeins}
\usepackage{tablefootnote}
\usepackage{algorithmic}
\usepackage[ruled,vlined]{algorithm2e}
\usepackage{paralist}
\usepackage{enumitem}
\usepackage[export]{adjustbox}

\usepackage[pagebackref=true,breaklinks=true,colorlinks,bookmarks=false]{hyperref}

\def\cvprPaperID{4384} 
\def\confYear{CVPR 2021}

\DeclareMathOperator*{\argmax}{arg\,max} 

\newcommand*{\blarrow}{\rotatebox[origin=c]{270}{$\Rsh$}}

\makeatletter
\newcommand\footnoteref[1]{\protected@xdef\@thefnmark{\ref{#1}}\@footnotemark}
\makeatother

\begin{document}

\title{Three Ways to Improve Semantic Segmentation\\ with Self-Supervised Depth Estimation}

\author{Lukas Hoyer\\
ETH Zurich\\
{\tt\small lhoyer@student.ethz.ch}
\and
Dengxin Dai\\
ETH Zurich\\
{\tt\small dai@vision.ee.ethz.ch}
\and
Yuhua Chen\\
ETH Zurich\\
{\tt\small yuhua.chen@vision.ee.ethz.ch}
\and
Adrian Köring\\
University of Bonn\\
{\tt\small adrian.koering@uni-bonn.de}
\and
Suman Saha\\
ETH Zurich \\
{\tt\small suman.saha@vision.ee.ethz.ch}
\and
Luc Van Gool\\
ETH Zurich \& KU Leuven\\
{\tt\small vangool@vision.ee.ethz.ch}
}

\maketitle

\begin{abstract}

Training deep networks for semantic segmentation requires large amounts of labeled training data, which presents a major challenge in practice, as labeling segmentation masks is a highly labor-intensive process. To address this issue, we present a framework for semi-supervised semantic segmentation, which is enhanced by self-supervised monocular depth estimation from unlabeled image sequences. 
In particular, we propose three key contributions: (1) We transfer knowledge from features learned during self-supervised depth estimation to semantic segmentation, (2) we implement a strong data augmentation by blending images and labels using the geometry of the scene, and (3) we utilize the depth feature diversity as well as the level of difficulty of learning depth in a student-teacher framework to select the most useful samples to be annotated for semantic segmentation.
We validate the proposed model on the Cityscapes dataset, where all three modules demonstrate significant performance gains, and we achieve state-of-the-art results for semi-supervised semantic segmentation. The implementation is available at \footnotesize{\url{https://github.com/lhoyer/improving_segmentation_with_selfsupervised_depth}}.
\end{abstract}

\section{Introduction}
\label{sec:introduction}

Convolutional Neural Networks (CNNs)~\cite{lecun1998gradient} have achieved state-of-the-art results for various computer vision tasks including semantic segmentation~\cite{long2015fully, chen2017deeplab}. However, training CNNs typically requires large-scale annotated datasets, due to millions of learnable parameters involved. Collecting such training data relies primarily on manual annotation. For semantic segmentation, the process can be particularly costly, due to the required dense annotations. For example, annotating a single image in the Cityscapes dataset took on average 1.5 hours~\cite{cordts2016cityscapes}.

Recently, self-supervised learning has shown to be a promising replacement for manually labeled data. It aims to learn representations from the structure of unlabeled data, instead of relying on a supervised loss, which involves manual labels. The principle has been successfully applied in depth estimation for stereo pairs~\cite{godard2017unsupervised} or image sequences~\cite{zhou2017unsupervised}. 
Additionally, semantic segmentation is known to be tightly coupled with depth. Several works have reported that jointly learning segmentation and \textit{supervised} depth estimation can benefit the performance of both tasks~\cite{vandenhende2020revisiting}. 
Motivated by these observations, we investigate the question: \textit{How can we leverage self-supervised depth estimation to improve semantic segmentation?} 

In this work, we propose a threefold approach to utilize self-supervised monocular depth estimation (SDE) 
\cite{godard2017unsupervised, zhou2017unsupervised, godard2019digging} 
to improve the performance of semantic segmentation and to reduce the amount of annotation needed.
Our contributions span across the holistic learning process from data selection, over data augmentation, up to cross-task representation learning, while being unified by the use of SDE.

First, we employ SDE as an auxiliary task for semantic image segmentation under a transfer learning and multi-task learning framework and show that it  noticeably improves the performance of semantic segmentation, especially when supervision is limited. Previous works only cover full supervision~\cite{klingner2020self}, pretraining~\cite{jiang2018self}, or improving SDE instead of segmentation~\cite{guizilini2020semantically}.
Second, we propose a strong data augmentation strategy, \emph{DepthMix}, which blends images as well as their labels according to the geometry of the scenes obtained from SDE. In comparison to previous methods~\cite{yun2019cutmix, olsson2020classmix}, DepthMix explicitly respects the geometric structure of the scenes and generates fewer artifacts (see Fig.~\ref{fig:depthmix}). 
And third, we propose an \emph{Automatic Data Selection for Annotation}, which selects the most useful samples to be annotated in order to maximize the gain. The selection is iteratively driven by two criteria: \emph{diversity} and \emph{uncertainty}. Both of them are conducted by a novel use of SDE as proxy task in this context.
While our method follows the active learning cycle (model training $\rightarrow$ query selection $\rightarrow$ annotation $\rightarrow$ model training)~\cite{settles2009active, yang2017suggestive}, it does not require a human in the loop to provide semantic segmentation labels as the human is replaced by a proxy-task SDE oracle. This greatly improves flexibility, scalability, and efficiency, especially considering crowdsourcing platforms for annotation.

The main advantage of our method is that we can learn from a large base of easily accessible unlabeled image sequences and utilize the learned knowledge to improve semantic segmentation performance in various ways. In our experimental evaluation on Cityscapes~\cite{cordts2016cityscapes}, we demonstrate significant performance gains of all three components and improve the previous state-of-the-art for semi-supervised segmentation by a considerable margin. Specifically, our method achieves 92\% of the full annotation baseline performance with only 1/30 available labels and even slightly outperforms it with only 1/8 labels.
Our contributions summarize as follows:
\begin{itemize}[topsep=4pt,itemsep=4pt, parsep=0pt]
    \item[(1)] To the best of our knowledge, we are the first to utilize SDE as an auxiliary task to exploit unlabeled image sequences and significantly improve the performance of semi-supervised semantic segmentation.
    \item[(2)] We propose \emph{DepthMix}, a strong data augmentation strategy, which respects the geometry of the scene and achieves, in combination with (1), state-of-the-art results for semi-supervised semantic segmentation.
    \item[(3)] We propose a novel \emph{Automatic Data Selection for Annotation} based on SDE to improve the flexibility of active learning. It replaces the human annotator with an SDE oracle and lifts the requirement of having a human in the loop of data selection.  
\end{itemize}
\section{Related Work}
\label{sec:related_work}

\subsection{(Semi-Supervised) Semantic Segmentation}

Since Convolutional Neural Networks (CNNs)~\cite{lecun1998gradient} were first used by Long \etal~\cite{long2015fully} for semantic segmentation, they have become the state-of-the-art method for this problem. Most architectures are based on an encoder decoder design such as \cite{long2015fully, ronneberger2015u, chen2018encoder}. Skip connections~\cite{ronneberger2015u} and dilated convolutions \cite{chen2014semantic, yu2015multi} preserve details in the segmentation and spatial pyramid pooling \cite{ghiasi2016laplacian, zhao2017pyramid, chen2017deeplab} aggregates different scales to exploit spatial context information. 

Semi-supervised semantic segmentation makes use of additional unlabeled data during training. For that purpose, Souly \etal~\cite{souly2017semi} and Hung \etal~\cite{hung2018adversarial} utilize generative adversarial networks \cite{goodfellow2014generative}. Souly \etal~\cite{souly2017semi} use that concept to generate additional training samples, while Hung \etal~\cite{hung2018adversarial} train the discriminator based on the semantic segmentation probability maps. s4GAN~\cite{mittal2019semi} extends this idea by adding a multi-label classification mean teacher~\cite{tarvainen2017mean}.
Another line of work~\cite{ouali2020semi, french2019consistency, olsson2020classmix} is based on consistency training, where perturbations are applied to unlabeled images or their intermediate features and a loss term enforces consistency of the segmentation. While Ouali \etal~\cite{ouali2020semi} study perturbation of encoder features, CutMix~\cite{french2019consistency} mixes crops from the input images and their pseudo-labels to generate additional training data, and ClassMix~\cite{olsson2020classmix} uses pseudo-label~\cite{lee2013pseudo} class segments to build the mix mask. 
Our proposed DepthMix module is inspired by these methods but, in contrast, it also respects the structure of the scene when mixing samples.
Commonly, several approaches~\cite{mittal2019semi, french2019consistency, olsson2020classmix, feng2020semi} include self-training with pseudo-labels~\cite{lee2013pseudo} and a mean teacher framework~\cite{tarvainen2017mean}, which is extended by Feng \etal~\cite{feng2020semi} with a class-balanced curriculum. Another related line of work is learning useful representations for semantic segmentation from self-supervised tasks such as tracking~\cite{wang2015unsupervised}, context inpainting~\cite{pathak2016context}, colorization~\cite{larsson2017colorization}, depth estimation~\cite{jiang2018self} (see Section~\ref{sec:segmentation_and_depth}), or optical flow prediction \cite{lee2019visuomotor}. However, all of these approaches are outperformed by ImageNet pretraining and are, therefore, not relevant for semi-supervised semantic segmentation in practice.

\subsection{Active Learning}

Another approach to reduce the number of required annotations is active learning. It iteratively requests the most informative samples to be labeled by a human. 
On the one side, uncertainty-based approaches select samples with a high uncertainty estimated based on, e.g., entropy~\cite{hwa2004sample, settles2008analysis} or ensemble disagreement~\cite{seung1992query, mccallumzy1998employing}. 
On the other side, diversity-based approaches select samples, which most increase the diversity of the labeled set~\cite{nguyen2004active, sener2017active, sinha2019variational}.
For segmentation, active learning is typically based on uncertainty measures such as MC dropout~\cite{gal2016dropout, yang2017suggestive, mackowiak2018cereals},
entropy~\cite{kasarla2019region, xie2020deal}, or multi-view consistency~\cite{siddiqui2020viewal}. In addition to methods selecting whole images~\cite{gorriz2017cost, yang2017suggestive, xie2020deal}, several approaches apply a more fine-grained label request at region level~\cite{mackowiak2018cereals, kasarla2019region, siddiqui2020viewal} and also include a label cost estimate~\cite{mackowiak2018cereals, kasarla2019region}. 

In contrast to these works, we perform automatic data selection for annotation by replacing the human with SDE as oracle. Therefore, we do not require human-in-the-loop annotation during the active learning cycle. Previous works performing unsupervised data selection are restricted to shallow models~\cite{yu2006active, zhang2011active, nie2013early, hu2013active, shi2016diversifying, li2018joint}, classification with low-dimensional inputs~\cite{li2020deep}, or do not perform an iterative data selection~\cite{zheng2019biomedical} to dynamically adapt to the uncertainty of the model trained on the currently labeled set.

\subsection{Improving Segmentation with SDE}
\label{sec:segmentation_and_depth}

Self-supervised depth estimation (SDE) aims to learn depth estimation from the geometric relations of stereo image pairs~\cite{garg2016unsupervised,godard2017unsupervised} or monocular videos~\cite{zhou2017unsupervised}. Due to the better availability of videos, we use the latter approach, where a neural network estimates depth and camera motion of two subsequent images and a photometric loss is computed after a differentiable warping. The approach has been improved by several follow-up works~\cite{godard2019digging,chen2019self,zou2018df}.

The combination of semantic segmentation and SDE was studied in previous works with the goal of improving \textit{depth} estimation. 
While \cite{ramirez2018geometry, jiao2018look, chen2019towards, klingner2020self} learn both tasks jointly, \cite{casser2019depth, guizilini2020semantically, jiang2019sense} distill knowledge from a teacher semantic segmentation network to guide SDE. 
To further utilize coherence between semantic segmentation and SDE, \cite{ramirez2018geometry, chen2019towards} proposed additional loss terms that encourage spatial proximity between depth discontinuities and segmentation contours. 

In contrast to these works, we do not aim to improve SDE but rather semi-supervised semantic segmentation. 
The closest to our approach are \cite{jiang2018self}, \cite{novoselboosting}, and \cite{klingner2020self}. 
Jiang \etal~\cite{jiang2018self} utilizes relative depth computed from optical flow to replace ImageNet pretraining for semantic segmentation.
In contrast, we additionally study multi-task learning of SDE and semantic segmentation and show that combining SDE with ImageNet features can even further boost performance.
Novosel \etal~\cite{novoselboosting} and Klingner \etal~\cite{klingner2020self} improve the semantic segmentation performance by jointly learning SDE. However, they focus on the fully-supervised setting, while our work explicitly addresses the challenges of semi-supervised semantic segmentation by using the depth estimates to generate additional training data and an automatic data selection mechanism based on SDE.
Another work supporting the usefulness of SDE for semantic segmentation from another viewpoint is \cite{klingner2020improved} demonstrating an improved noise and attack robustness.

\section{Methods}
\label{sec:methods}

\begin{figure}[tb]
\centering
\includegraphics[width=0.85\linewidth]{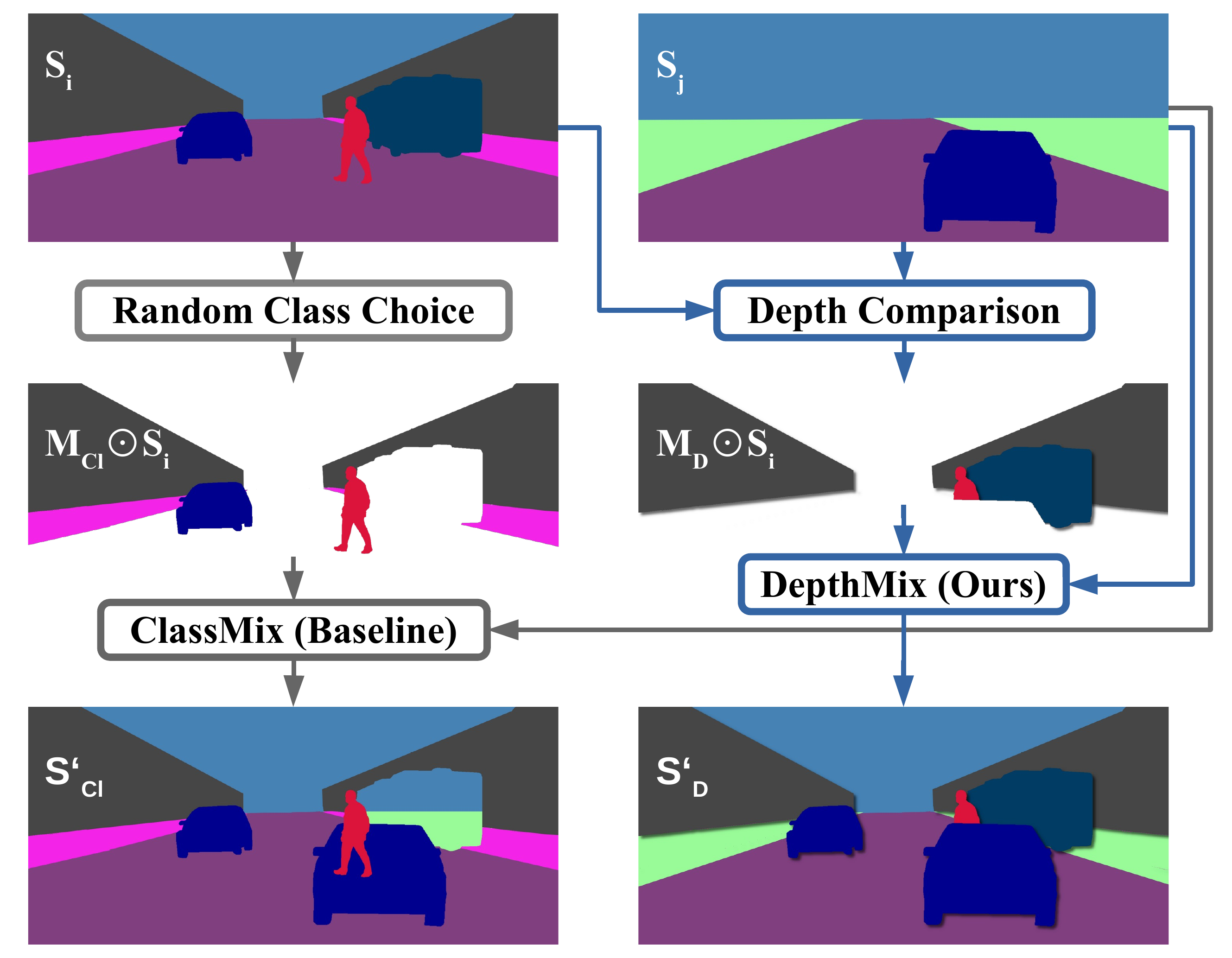}
\caption{Concept of the proposed DepthMix augmentation (refer to Sec.~\ref{sec:depthmix}) and its baseline ClassMix~\cite{olsson2020classmix}. By utilizing SDE, DepthMix mitigates geometric artifacts.}
\label{fig:depthmix}
\end{figure}

In this section, we present our three ways to improve the performance of semantic segmentation with self-supervised depth estimation (SDE). They focus on three different aspects of semantic segmentation, covering data selection for annotation, data augmentation, and multi-task learning. Given $N$ images and $K$ image sequences from the same domain, our first method, \emph{Automatic Data Selection for Annotation}, uses SDE learned on the $K$ (unlabeled) sequences to select $N_A$ images out of the $N$ images for human annotation (see Alg.~\ref{alg:label_selection}). Our second approach, termed \emph{DepthMix}, leverages the learned SDE to create geometrically-sound `virtual' training samples from pairs of labeled images and their annotations (see Fig.~\ref{fig:depthmix}). Our third method learns semantic segmentation with SDE as an auxiliary task under a multi-tasking framework (see Fig.~\ref{fig:architecture}). The learning is reinforced by a multi-task pretraining process combining SDE with image classification. 

For SDE, we follow the method of Godard \etal~\cite{godard2019digging}, which we briefly introduce in the following. We first train a depth estimation network $f_D$ to predict the depth of a target image and a pose estimation network $f_T$ to estimate the camera motion from the target image and the source image. Depth and pose are used to produce a differentiable warping to transform the source image into the target image. The photometric error between the target image and multiple warped source frames is combined by a pixel-wise minimum. Besides, stationary pixels are masked out and an edge-aware depth smoothness term is applied resulting in the final self-supervised depth loss $L_D$. We refer the reader to the original paper~\cite{godard2019digging} for more details.

\subsection{Automatic Data Selection for Annotation}

We use SDE as proxy task for selecting $N_A$ samples out of a set of $N$ unlabeled samples for a human to create semantic segmentation labels.
The selection is conducted progressively in multiple steps, similar to the standard active learning cycle (model training $\rightarrow$ query selection $\rightarrow$ annotation $\rightarrow$ model training). However, our data selection is fully automatic and does not require a human in the loop as the annotation is done by a proxy-task SDE oracle.

Let's denote by $\mathcal{G}$, $\mathcal{G}_A$, and $\mathcal{G}_U$, the whole image set, the selected sub-set for annotation, and the un-selected sub-set.   
Initially, we have $\mathcal{G}_A=\emptyset$ and $\mathcal{G}_U=\mathcal{G}$. The selection is driven by two criteria: \emph{diversity} and \emph{uncertainty}. Diversity sampling encourages that selected images are diverse and cover different scenes. Uncertainty sampling favors adding unlabeled images that are near a decision boundary (with high uncertainties) of the model trained on the current $\mathcal{G}_A$. 
For uncertainty sampling, we need to train and update the model with $\mathcal{G}_A$. It is inefficient to repeat this every time a new image is added. For the sake of efficiency, we divide the selection into $T$ steps and only train the model $T$ times. In each step $t$, $n_t$ images are selected and moved from $\mathcal{G}_U$ to $\mathcal{G}_A$, so we have $\sum_{t=1}^T n_t = N_A$. After each step $t$, a model is trained on $\mathcal{G}_A$ and evaluated on $\mathcal{G}_U$ to get updated uncertainties for step $t+1$.

\begin{algorithm}[tb]
\caption{Automatic Data Selection}
\label{alg:label_selection}
\begin{algorithmic}[1]
\STATE $t=1$
\STATE $i \gets \text{uniform}(1,N)$ 
\STATE $\mathcal{G}_A = \{I_i\} \text{ and } \mathcal{G}_U = \mathcal{G}_U \setminus \{I_i\}$ 
\FOR{$k=2$ {\bfseries to} $N_A$}
\IF{ $k==\sum_{t'=1}^tn_{t'}$}
\STATE Train depth student $\Phi_\text{SIDE}$ on the current $\mathcal{G}_A$
\STATE Calculate $E(i) \text{ } \forall I_i \in \mathcal{G}_U$
\STATE $t = t + 1$
 \ENDIF
 \IF{$t==1$} 
     \STATE Obtain index $i$ according to Eq. \ref{eq:diversity} 
 \ELSE 
      \STATE Obtain index $i$ according to Eq. \ref{eq:uncertainty:diversity} 
 \ENDIF
    \STATE $\mathcal{G}_A = \mathcal{G}_A \cup \{I_i\} \text{ and } \mathcal{G}_U = \mathcal{G}_U \setminus \{I_i\}$
\ENDFOR
\end{algorithmic}
\end{algorithm}

\noindent\textbf{Diversity Sampling}:
To ensure that the chosen annotated samples are diverse enough to represent the entire dataset well, we use an iterative farthest point sampling based on the L2 distance over features $\Phi^{\text{SDE}}$ computed by an intermediate layer of the SDE network.
At step $t$, for each of the $n_t$ samples, we choose the one in $\mathcal{G}_U$ with the largest distance to the current annotation set $\mathcal{G}_A$.
The set of selected samples $\mathcal{G}_A$ is iteratively extended by moving one image at a time from $\mathcal{G}_U$ to $\mathcal{G}_A$ until the $n_t$ images are collected:
\begin{equation}
\label{eq:set:enlarge}
    \mathcal{G}_U  = \mathcal{G}_U  \setminus \{I_i\} \text{ and } \mathcal{G}_A =  \mathcal{G}_A \cup \{I_i\} ,
\end{equation}
\begin{equation}
\label{eq:diversity}
     i = \argmax_{I_i\in \mathcal{G}_U} \min_{I_j\in \mathcal{G}_A} ||\Phi^{\text{SDE}}_i - \Phi^\text{SDE}_j||_2.
\end{equation}

\noindent\textbf{Uncertainty Sampling}:
While Diversity Sampling is able to select diverse new samples, it is unaware of the uncertainties of a semantic segmentation model over these samples. Uncertainty Sampling aims to select difficult samples, \ie, samples in $\mathcal{G}_U$ that the model trained on the current $\mathcal{G}_A$ cannot handle well.
In order to train this model, active learning typically uses a human-in-the-loop strategy to add annotations for selected samples. In this work, we use a proxy task based on self-supervised annotations, which can run automatically, to make the method more flexible and efficient.
Since our target task is single-image semantic segmentation, we choose to use single-image depth estimation (SIDE) as the proxy task. Importantly, due to our SDE framework, depth pseudo-labels are available for $\mathcal{G}$. Using these pseudo-labels, we train a SIDE method on $\mathcal{G}_A$ and measure the uncertainty of its depth predictions on $\mathcal{G}_U$. Due to the high correlation of single-image semantic segmentation and SIDE, the generated uncertainties are informative and can be used to guide our sampling procedure. As the depth student model is trained only on $\mathcal{G}_A$, it can specifically approximate the difficulty of candidate samples with respect to the already selected samples in $\mathcal{G}_A$. The student is trained from scratch in each step $t$, instead of being fine-tuned from $t-1$, to avoid getting stuck in the previous local minimum. Note that the SDE method is trained on a much larger unlabeled dataset, \ie, the $K$ image sequences, and can provide good guidance for the SIDE method.

In particular, the uncertainty is signaled by the disparity error between the student network $f_{\text{SIDE}}$ and the teacher network $f_{\text{SDE}}$ in the log-scale space under L1 distance: 
\begin{equation}
\label{eq:uncertainty}
    E(i) = || \log(1 + f_{\text{SDE}}(I_i)) - \log(1 + f_{\text{SIDE}}(I_i))||_1.
\end{equation}
As the disparity difference of far-away objects is small, the log-scale is used to avoid the loss being dominated by close-range objects. This criterion can be added into Eq. \ref{eq:diversity} to also select samples with higher uncertainties for the dataset update in Eq. \ref{eq:set:enlarge}: 
\begin{equation}
    \label{eq:uncertainty:diversity}
     i = \argmax_{I_i\in \mathcal{G}_U} \min_{I_j\in \mathcal{G}_A} ||\Phi^{\text{SDE}}_i - \Phi^\text{SDE}_j||_2 + \lambda_{\text{E}}E(i),
\end{equation}
where $\lambda_E$ is a parameter to balance the contribution of the two terms. For diversity sampling, we still use SDE features instead of SIDE student features as SDE is trained on the entire dataset, which provides better features for diversity estimation.
When $n_t$ images have been selected according to Eq. \ref{eq:set:enlarge} and Eq. \ref{eq:uncertainty:diversity} at step $t$, a new SIDE model will be trained on the current $\mathcal{G}_A$ in order to continue further.
As presented previously, our selection proceeds progressively in $T$ steps until we collect all $N_A$ images. The algorithm of this selection is summarized in Alg. \ref{alg:label_selection}, where $\sum_{t'=1}^tn_{t'}$ describes the desired size of $\mathcal{G}_A$ at the end of step $t$.

\subsection{DepthMix Data Augmentation}
\label{sec:depthmix}

Inspired by the recent success of data augmentation approaches that mixup pairs of images and their (pseudo) labels to generate more training samples for semantic segmentation~\cite{yun2019cutmix, french2019consistency, olsson2020classmix}, we propose an algorithm, termed DepthMix, to utilize self-supervised depth estimates to maintain the integrity of the scene structure during mixing.

Given two images $I_i$ and $I_j$ of the same size, we would like to copy some regions from $I_i$ and paste them directly into $I_j$ to get a virtual sample $I^\prime$. The copied regions are indicated by a mask $M$, which is a binary image of the same size as the two images. The image creation is done as
\begin{equation}
    I^\prime = M \odot I_i + (1 - M) \odot I_j,
    \label{eq:mix}
\end{equation}
where $\odot$ denotes the element-wise product. 
The label maps of the two images $S_i$ and $S_j$ are mixed up with the same mask $M$ to generate $S^\prime$. The mixing can be applied to labeled data and unlabeled data using human ground truths or pseudo-labels, respectively.
Existing methods generate this mask $M$ in different ways, \eg, randomly sampled rectangular regions \cite{yun2019cutmix, french2019consistency} or randomly selected object segments \cite{olsson2020classmix}.
In those methods, the structure of the scene is not considered and foreground and background are not distinguished.
We find images synthesized by these methods often violate the geometric relationships between objects. For instance, a distant object can be copied onto a close-range object or only unoccluded parts of mid-range objects are copied onto the other image. Imagine how strange it is to see a pedestrian standing on top of a car or to see sky through a hole in a building (just as shown in Fig.~\ref{fig:depthmix} left).

Our DepthMix is designed to mitigate this issue. It uses the estimated depth $\hat{D}i$ and $\hat{D}j$ of the two images to generate the mix mask $M$ that respects the notion of geometry. It is implemented by selecting only pixels from $I_i$ whose depth values are smaller than the depth values of the pixels at the same locations in $I_j$: 
\begin{equation}
    M(a,b) = \left\{ 
    \begin{array}{rl} 
    1 & \text{if } \hat{D}_i(a,b) < \hat{D}_j(a,b) + \epsilon \\
    0 & \text{otherwise } 
     \end{array} \right.
\end{equation}
where $a$ and $b$ are pixel indices, and $\epsilon$ is a small value to avoid conflicts of objects that are naturally at the same depth plane such as road or sky.
By using this $M$, DepthMix respects the depth of objects in both images, such that only closer objects can occlude further-away objects. We illustrate this advantage of DepthMix with an example in Fig.~\ref{fig:depthmix}.

\begin{figure}
\centering
\includegraphics[width=1\linewidth]{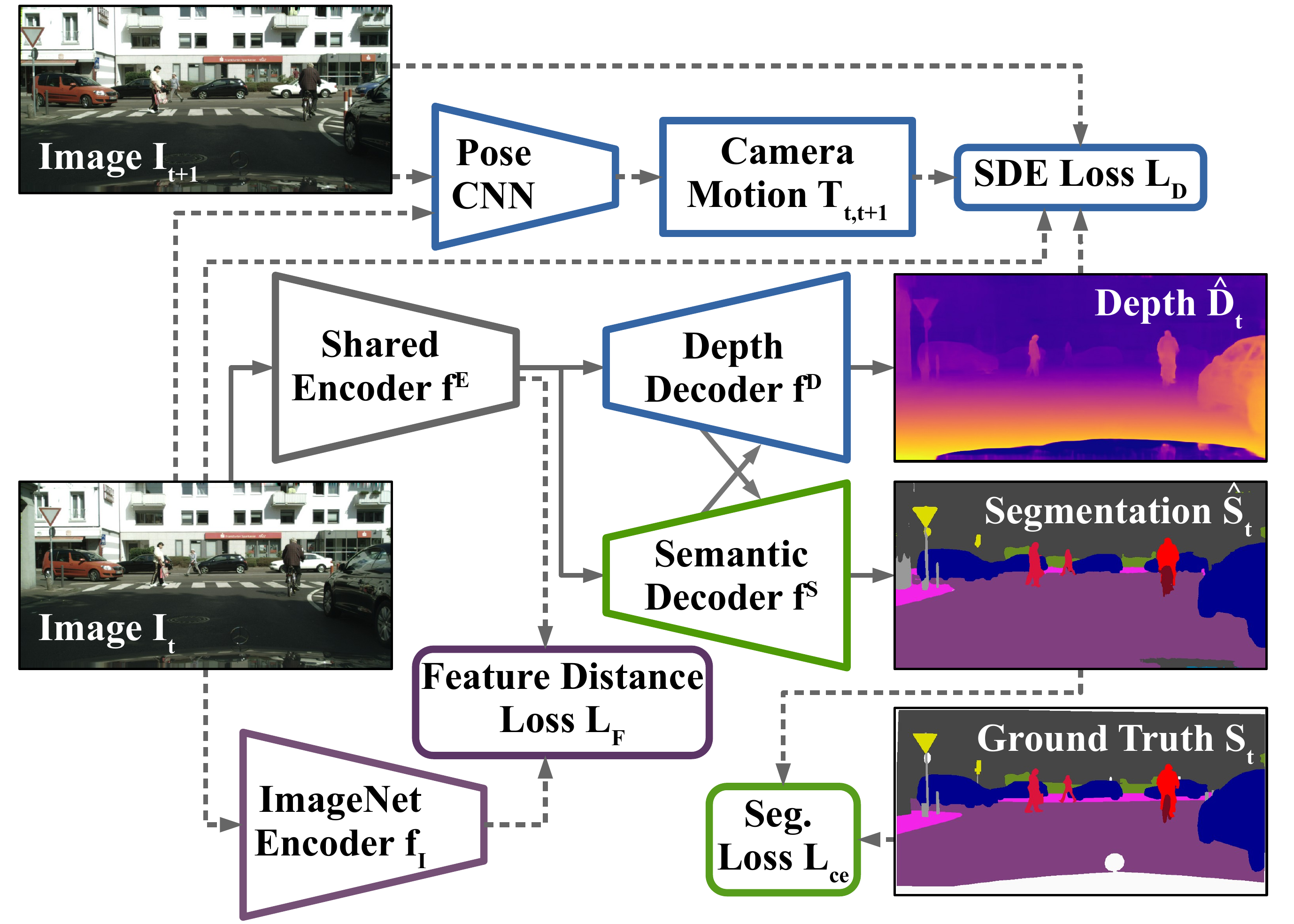}
\caption{Architecture for learning semantic segmentation with SDE as auxiliary task according to Sec.~\ref{sec:semisupseg}. The dashed paths are only used during training and only if image sequences and/or segmentation ground truth are available for a training sample.}
\label{fig:architecture}
\end{figure}

\subsection{Semi-Supervised Semantic Segmentation}
\label{sec:semisupseg}

In this section, we train a semantic segmentation model utilizing the labeled image dataset $\mathcal{G}_A$, the unlabeled image dataset $\mathcal{G}_U$, and $K$ unlabeled image sequences. We first discuss how to exploit SDE on the image sequences to improve our semantic segmentation. We then show how to use $\mathcal{G}_U$ to further improve the performance.

\noindent\textbf{Learning with Auxiliary Tasks}:
For learning semantic segmentation and SDE jointly, we use a network with shared encoder $f^E_\theta$ and a separate depth $f^D_\theta$ and segmentation decoder $f^S_\theta$ (see Fig.~\ref{fig:architecture}). The depth branch is trained using the SDE loss $L_D$ and the segmentation branch $g^S_\theta = f^S_\theta \circ f^E_\theta$ is trained using the pixel-wise cross-entropy $L_{ce}$.

In order to initialize the pose estimation network and the depth decoder properly, the architecture is first trained on $K$ unlabeled image sequences for SDE. As a common practice, we initialize the encoder with ImageNet weights as they provide useful semantic features learned during image classification. To avoid forgetting semantic features during the SDE pretraining, we utilize a feature distance loss between the current bottleneck features $f^{E}_\theta$ and the bottleneck features of the encoder with ImageNet weights $f^{E}_{I}$:
\begin{equation}
    L_{F} = ||f^{E}_\theta - f^{E}_{I}||_2.
\end{equation}
The loss for the depth pretraining is the weighted sum of the SDE loss and the ImageNet feature distance loss:
\begin{equation}
    L_P = L_D + \lambda_{F} L_{F}.
\end{equation}

To additionally incorporate transfer learning from depth estimation to semantic segmentation, the weights of $f^{D}_\theta$ are used to initialize $f^{S}_\theta$. 
For effective multi-task learning, we use an attention-guided distillation module~\cite{xu2018pad} to exchange useful intermediate features between both decoders.

\noindent\textbf{Learning with Unlabeled Images}:
In order to further utilize the unlabeled dataset $\mathcal{G}_U$, we generate pseudo-labels using the mean teacher algorithm~\cite{tarvainen2017mean}, which is commonly used in semi-supervised learning~\cite{berthelot2019mixmatch, verma2019interpolation, french2019consistency, olsson2020classmix}. For that purpose, an exponential moving average is applied to the weights of the semantic segmentation model $g^S_\theta$ to obtain the weights of the mean teacher $\theta_T$:
\begin{equation}
    \theta'_T = \alpha \theta_T + (1 - \alpha) \theta.
\end{equation}
To generate the pseudo-labels, an argmax over the classes $C$ is applied to the prediction of the mean teacher. 
\begin{equation}
    S_U = \argmax_{c \in C}(g^S_{\theta_T}(I_U)).
\end{equation}
The mean teacher can be considered as a temporal ensemble, resulting in stable predictions for the pseudo-labels, while the argmax ensures confident predictions~\cite{olsson2020classmix}. 

For the semi-supervised setting, the segmentation network is trained with labeled samples ($I_A$, $S_A$) and pseudo-labeled samples ($I_U$, $S_U$):
\begin{equation}
\begin{split}
    L_{SSL} = L_{ce}(g^S_\theta(I_A), S_A) +
    \lambda_P(S_U) L_{ce}(g^S_\theta(I_U), S_U)) 
\end{split}
\label{eq:L_SSL_1}
\end{equation}
$\lambda_P(S_U)$ is chosen to reflect the quality of the pseudo-label represented by the fraction of pixels exceeding a threshold $\tau$ for the predicted probability of the most confident class $\max_{c \in C}(g^S_{\theta_T}(I_U))$, as suggested in \cite{olsson2020classmix}. We incorporate DepthMix samples ($I'$, $S'$), which are obtained from the combined labeled and pseudo-labeled data pool $I_i, I_j \in \mathcal{G}_A \cup \mathcal{G}_U$ (see Eq. \ref{eq:mix}), into Eq.~\ref{eq:L_SSL_1} to replace the unlabeled samples ($S_U$, $L_U$). Our semi-supervised learning is now changed to:
\begin{equation}
\begin{split}
    L_{SSL} = L_{ce}(g^S_\theta(I_A), S_A) +
    \lambda_P(S') L_{ce}(g^S_\theta(I'), S')). 
\end{split}
\label{eq:L_SSL_2}
\end{equation}

\section{Experiments}
\label{sec:experiments}

\subsection{Implementation Details}

\noindent\textbf{Dataset}: 
We evaluate our method on the Cityscapes dataset~\cite{cordts2016cityscapes}, which consists of 2975 training and 500 validation images with semantic segmentation labels from European street scenes. We downsample the images to $1024 \times 512$ pixels. Besides, random cropping to a size of $512 \times 512$ and random horizontal flipping are used in the training. Importantly, Cityscapes provides 20 unlabeled frames before and 10 after the labeled image, which are used for SDE training. During the semi-supervised segmentation, only the originally 2975 labeled training images are used. They are randomly split into a labeled and an unlabeled subset. 

\noindent\textbf{Network Architecture}: 
Our network consists of a shared ResNet101~\cite{he2016deep} encoder with output stride 16 and a separate decoder for segmentation and SDE. The decoder consists of an ASPP~\cite{chen2017deeplab} block to aggregate features from multiple scales and another four upsampling blocks with skip connections~\cite{ronneberger2015u}. For SDE, the upsampling blocks have a disparity side output at the respective scale. For effective multi-task learning, we additionally follow PAD-Net~\cite{xu2018pad} and deploy an attention-guided distillation module after the third decoder block. It serves the purpose of exchanging useful features between segmentation and depth estimation. 

\noindent\textbf{Training}:
For the SDE pretraining, the depth and pose network are trained using Adam~\cite{kingma2014adam}, a batch size of 4, and an initial learning rate of $1 \times 10^{-4}$, which is divided by 10 after 160k iterations. The SDE loss is calculated on four scales with three subsequent images. During the first 300k iterations, only the depth decoder and the pose network are trained. Afterwards, the depth encoder is fine-tuned with an ImageNet feature distance $\lambda_F = 1 \times 10^{-2}$ for another 50k iterations. The encoder is initialized with ImageNet weights, either before depth pretraining or before semantic segmentation if depth pretraining is ablated.

For the multi-task setting, we train the network using SGD with a learning rate of $1 \times 10^{-3}$ for the encoder and depth decoder, $1 \times 10^{-2}$ for the segmentation decoder, and $1 \times 10^{-6}$ for the pose network. The learning rate is reduced by 10 after 30k iterations and trained for another 10k iterations. A momentum of 0.9, a weight decay of $5 \times 10^{-4}$, and a gradient norm clipping to 10 are used. The loss for segmentation and SDE are weighted equally. The mean teacher has $\alpha=0.99$ and within an iteration, the network is trained on a clean labeled and an augmented mixed batch with size 2, respectively. The latter uses DepthMix with $\epsilon = 0.03$, color jitter, and Gaussian blur.

\noindent\textbf{Data Selection for Annotation}:
In the data selection experiment, we use a slimmed network architecture with a ResNet50 encoder and fewer decoder channels for $f_{SIDE}$. It is trained using Adam with $1 \times 10^{-4}$ learning rate and polynomial decay with exponent 0.9 for faster convergence.
For calculating the depth feature diversity, we use the output of the second depth decoder block after SDE pretraining. It is downsampled by average pooling to a size of 8x4 pixels and the feature channels are normalized to zero-mean unit-variance over the dataset. The student depth error is weighted by $\lambda_E = 1000$. The number of the selected samples ($\sum_{t'=1}^t n_{t'}$) is iteratively increased to 25, 50, 100, 200, 372, and 744.
For each subset, a student depth network is trained from scratch for 4k, 8k, 12k, 16k, and 20k iterations, respectively, to calculate the student depth error.

\begin{table*}
\caption{Performance on the Cityscapes validation set (mIoU in \%, standard deviation over 3 random seeds).}
\label{tab:comp}
\vspace{0.2cm}
\centering
\setlength{\tabcolsep}{3pt}
\begin{tabular}{|l|ll|ll|ll|ll|}
\hline
Labeled Samples & 1/30 (100)    &        & 1/8 (372)     &       & 1/4 (744)     &       & Full (2975) &       \\\hline\hline
Baseline \cite{hung2018adversarial}        & --             &        & 55.50 & \blarrow      & 59.90 & \blarrow      & 66.40     & \blarrow      \\
Adversarial \cite{hung2018adversarial}      & --       &        & 58.80 & +3.30 & 62.30 & +2.40 & --           &       \\\hline\hline
Baseline \cite{mittal2019semi}        & --             &        & 56.20 & \blarrow      & 60.20 & \blarrow      & 66.00     &      \\
s4GAN \cite{mittal2019semi}           & --       &        & 59.30 & +3.10 & 61.90 & +1.70 & 65.80     & --0.20 \\\hline\hline
Baseline \cite{french2019consistency}        & 44.41 \scriptsize{$\pm1.11$}       & \blarrow          & 55.25 \scriptsize{$\pm0.66$} & \blarrow      & 60.57 \scriptsize{$\pm1.13$} & \blarrow      & 67.53 \scriptsize{$\pm0.35$}     & \blarrow      \\
CutMix \cite{french2019consistency}         & 51.20 \scriptsize{$\pm2.29$} & +6.79  & 60.34 \scriptsize{$\pm1.24$} & +5.09 & 63.87 \scriptsize{$\pm0.71$} & +3.30 & 67.68 \scriptsize{$\pm0.37$}     & +0.15 \\\hline\hline
Baseline \cite{feng2020semi}        & 45.50       & \blarrow          & 56.70 & \blarrow      & 61.10 & \blarrow      & 66.90     &      \\
DST--CBC \cite{feng2020semi}     & 48.70   & +3.20  & 60.50 & +3.80 & 64.40 & +3.30 & --           &       \\\hline\hline
Baseline \cite{olsson2020classmix}        & 43.84 \scriptsize{$\pm0.71$}      & \blarrow       & 54.84 \scriptsize{$\pm1.14$} & \blarrow      & 60.08 \scriptsize{$\pm0.62$} & \blarrow      & 66.19 \scriptsize{$\pm0.11$}    &      \\
ClassMix \cite{olsson2020classmix}         & 54.07 \scriptsize{$\pm1.61$} & \underline{+10.23} & 61.35 \scriptsize{$\pm0.62$} & +6.51 & 63.63 \scriptsize{$\pm0.33$} & +3.55 & --           &       \\\hline\hline
Baseline & 48.75	\scriptsize{$\pm$1.61} & \blarrow &	59.14	\scriptsize{$\pm$1.02}	& \blarrow &	63.46	\scriptsize{$\pm$0.38}	& \blarrow &	67.77	\scriptsize{$\pm$0.13} & \blarrow \\
ClassMix \cite{olsson2020classmix}\tablefootnote{\label{fn:setting} Results of the reimplementation in our experiment setting.} & 56.82	\scriptsize{$\pm$1.65} &	+8.07 &	63.86	\scriptsize{$\pm$0.41} &	+4.72 &	65.57	\scriptsize{$\pm$0.71} &	+2.11 &	--		& \\
ClassMix~\cite{olsson2020classmix} (+Video) & 56.79	\scriptsize{$\pm$1.98} & +8.04 & 63.22 \scriptsize{$\pm$0.84} & +4.08 &	65.72 \scriptsize{$\pm$0.18} & +2.26 & \underline{68.23}	\scriptsize{$\pm$0.70}		& \underline{+0.46} \\
Ours	& \underline{58.40}	\scriptsize{$\pm$1.36} &	+9.65 &	\underline{66.66}	\scriptsize{$\pm$1.05} &	\underline{+7.52} &	\underline{68.43}	\scriptsize{$\pm$0.06} &	\underline{+4.98} &	\textbf{71.16}	\scriptsize{$\pm$0.16} &	\textbf{+3.40} \\
Ours (+Data Selection) & \textbf{62.09}	\scriptsize{$\pm$0.39} &	\textbf{+13.34} &	\textbf{68.01}	\scriptsize{$\pm$0.83} &	\textbf{+8.87} &	\textbf{69.38}	\scriptsize{$\pm$0.33} &	\textbf{+5.92} & -- & \\
\hline
\end{tabular}
\end{table*}

\subsection{Semi-Supervised Semantic Segmentation}

\begin{figure}
    \centering
    \includegraphics[width=0.9\linewidth]{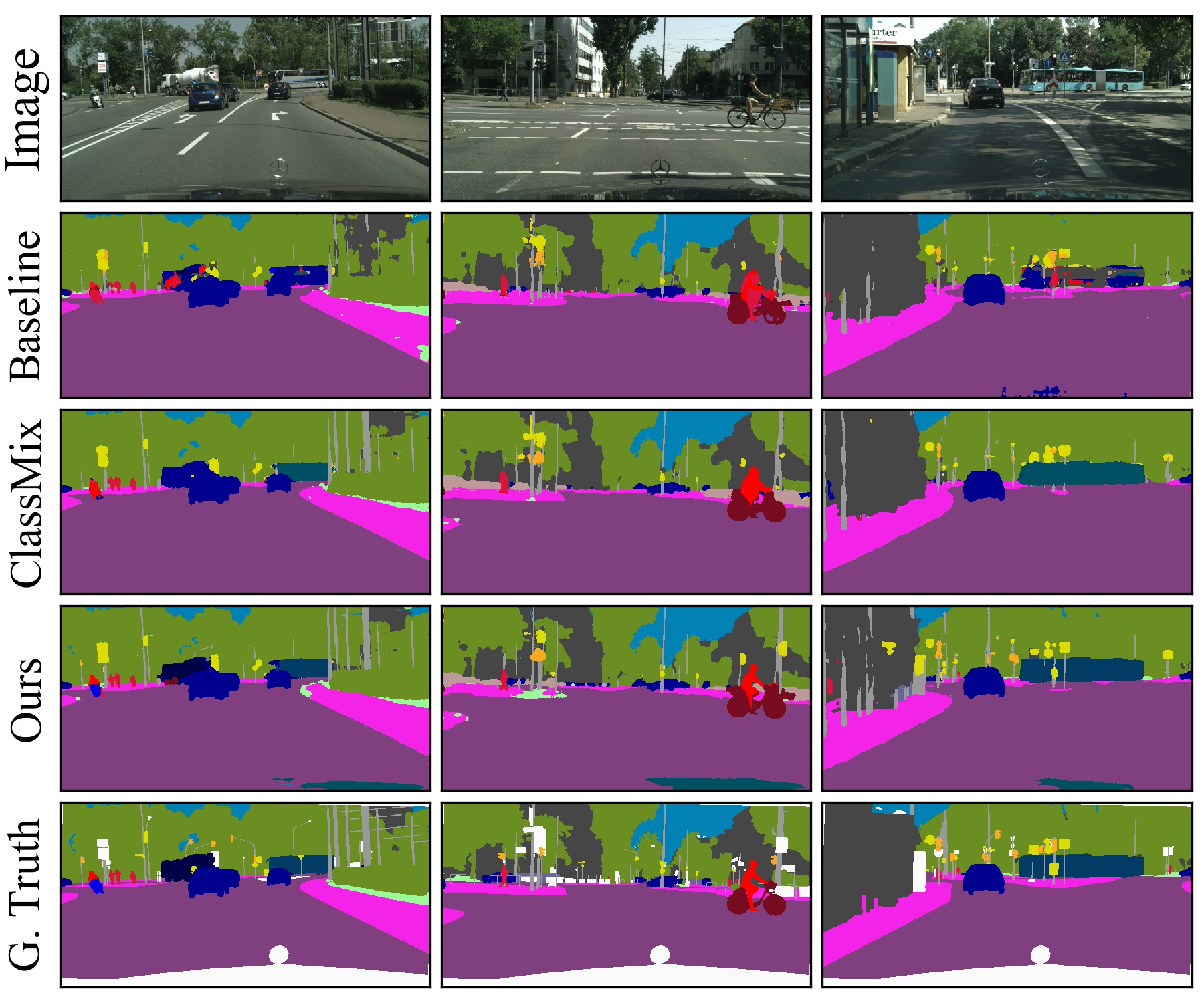}
    \caption{Example semantic segmentations of our method for 100 labeled samples in comparison with ClassMix~\cite{olsson2020classmix}.}
    \label{fig:examples}
\end{figure}

First, we compare our approach with several state-of-the-art semi-supervised learning approaches. We summarize the results in Tab.~\ref{tab:comp}. The performance (mIoU in \%) of the semi-supervised methods and their baselines (only trained on the labeled dataset) are shown for a different number of labeled samples. As the performance of the baselines differs, there are columns showing the absolute improvement for better comparability. As our baseline utilizes a more capable network architecture due to the U-Net decoder with ASPP as opposed to a DeepLabv2 decoder used by most previous works, we also reimplemented the state-of-the-art method, ClassMix~\cite{olsson2020classmix} with our network architecture and training parameters to ensure a direct comparison.

As shown in Tab.~\ref{tab:comp}, our method (without data selection) outperforms all other approaches on each labeled subset size for both the absolute performance as well as the improvement to the baseline. The only exception is the absolute improvement of the original results of ClassMix for 100 labeled samples. However, if we consider ClassMix trained in our setting, our method outperforms it also in this case. This can be explained by the considerably higher baseline performance in our setting, which increases the difficulty to achieve an high improvement. 
Adding data selection even further increases the performance by a significant margin, so that our method, trained with only 1/8 of the labels, even slightly outperforms the fully-supervised baseline.

To identify whether the improvement originates from access to more unlabeled data or from the effectiveness of our approach, we compare to another baseline ``ClassMix (+Video)". More specifically, we also provide all unlabeled image sequences to ClassMix and see how much it can benefit from this additional amount of unlabeled data. Experimental results show no significant difference. This is probably due to the high correlation of the Cityscapes image dataset and the video dataset (the images are the 20th frames of the video clips).

The adequacy of our approach is also reflected in the example predictions in Fig.~\ref{fig:examples}. We can observe that the contours of classes are more precise. Moreover, difficult objects such as bus, train, rider, or truck can be better distinguished. This observation is also quantitatively confirmed by the class-wise IoU improvement shown in Fig.~\ref{fig:classwise_improvement}.

\subsection{Ablation Study}

\begin{table}
\caption{Ablation of the architecture components (D-T: SDE Transfer Learning, D-M SDE Transfer and Multi-Task Learning, F: ImageNet Feature Distance Loss, P: Pseudo-Labeling, X-C: Mix Class, X-D: Mix Depth, S - Data Selection). mIoU in \%, standard deviation over 3 seeds.}
\label{tab:ablation}
\vspace{0.2cm}
\centering
\setlength{\tabcolsep}{2.5pt}
\begin{tabular}{cccccllll}
 \hline
D & F & P & X & S & \multicolumn{2}{l}{372 Samples} & \multicolumn{2}{l}{2975 Samples} \\
\hline\hline
  &     &    &   &     & 59.14 \scriptsize{$\pm$1.02}  & \blarrow  & 67.77 \scriptsize{$\pm$0.13}  & \blarrow\\
T &     &    &   &     & 60.46 \scriptsize{$\pm$0.64}  & +1.31     & 69.00 \scriptsize{$\pm$0.70}  & +1.23 \\
T & \checkmark   &    &   &     & 60.80 \scriptsize{$\pm$0.69}  & +1.66     & 69.47 \scriptsize{$\pm$0.38}  & +1.71 \\
M & \checkmark   &    &   &     & 61.25 \scriptsize{$\pm$0.55}  & +2.10     & 69.76 \scriptsize{$\pm$0.39}  & +1.99 \\
\hline
  &     & \checkmark  &   &     & 62.39 \scriptsize{$\pm$0.86}  & +3.24     & --                             &       \\
  &     & \checkmark  & C  &     & 63.16	\scriptsize{$\pm$0.89}   & +4.02    &   69.60	\scriptsize{$\pm$0.32} & +1.83 \\
  &     & \checkmark  & D  &     & 64.14	\scriptsize{$\pm$1.34} & +5.00 & 69.83	\scriptsize{$\pm$0.36} & +2.06 \\
M & \checkmark   & \checkmark  & D &     & 66.66 \scriptsize{$\pm$1.05}  &	+7.52     & 71.16 \scriptsize{$\pm$0.16}  & +3.40 \\
\hline
  &              &             &     & \checkmark   & 64.25 \scriptsize{$\pm$ 0.18} &    +5.11 & -- & \\
M & \checkmark   & \checkmark  & D & \checkmark   & 68.01	\scriptsize{$\pm$0.83} &	+8.87 & -- & \\
\hline
\end{tabular}
\end{table}

\begin{figure}
    \centering
    \includegraphics[width=1.0\linewidth]{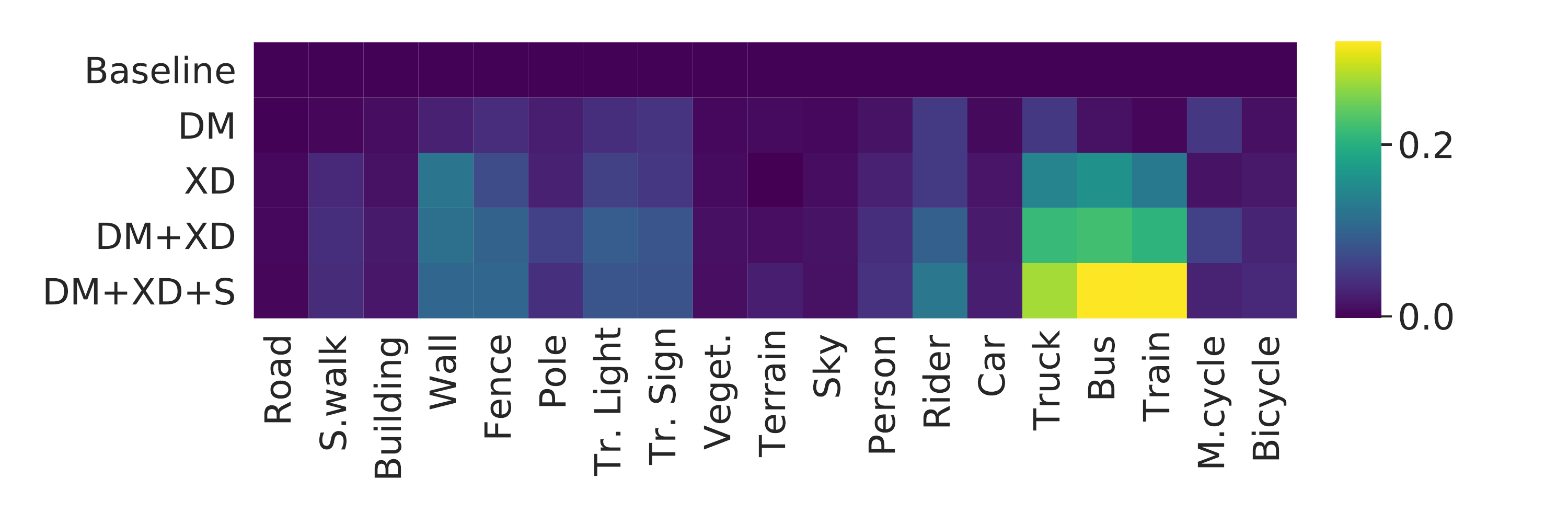}
    \caption{Improvement of the class-wise IoU over the baseline performance for 372 labeled samples (DM: SDE Multi-Task Learning, XD: DepthMix with Pseudo-Labels, S:~Data Selection).}
    \label{fig:classwise_improvement}
\end{figure}

\begin{figure}
    \centering
    \includegraphics[width=1.0\linewidth]{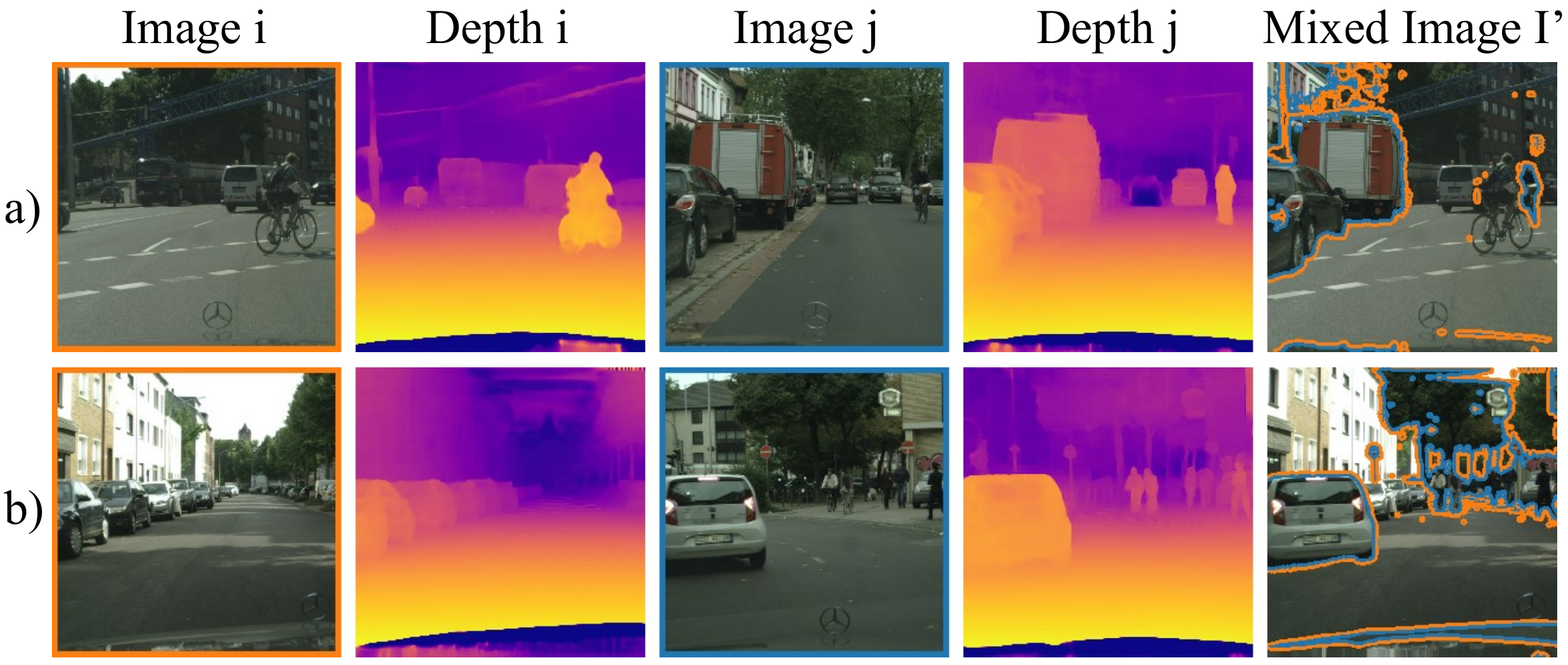}
    \caption{DepthMix applied to Cityscapes crops.}
    \label{fig:depthmix_paper_examples}
\end{figure}

Next, we analyze the individual contribution of each component of the proposed method. For this purpose, we test several ablated versions of our model for both the cases of 372 and 2975 labeled samples. We summarize the results in Tab.~\ref{tab:ablation}. 
It can be seen that each contribution adds a significant performance improvement over the baseline. For 372 (2975) annotated samples, transfer and multi-task learning improve the performance by +2.10 (+1.99), DepthMix with pseudo-labels by +5.00 (+2.06), and automatic data selection by +5.11 (--) mIoU percentage points. As our components are orthogonal, combining them even further increases performance. SDE Multi-Tasking and DepthMix achieve +7.52 (+3.40) and all three components +8.87 (--) mIoU percentage points improvement.
Note that the high variance for few labeled samples is mostly due to the high influence of the randomly selected labeled subset. The chosen subset affects all configurations equally and the reported improvements are consistent for each subset.

Furthermore, we compare DepthMix with ClassMix as a standalone. For a fair comparison, we additionally include mixing labeled samples with their ground truth to ClassMix. It can be seen that DepthMix outperforms the ClassMix by 0.98 (0.23) percentage points for 372 (2975) annotated samples, which shows the effect of the geometry aware augmentation. Fig.~\ref{fig:depthmix_paper_examples} shows DepthMix examples demonstrating that SDE allows to correctly model occlusions and to produce synthetic samples with a realistic appearance.

For more insights into possible reasons for these improvements, we visualize the improvement of the architecture components over the baseline for each class separately in Fig.~\ref{fig:classwise_improvement}. It can be seen that depth multi-task learning (DM) improves mostly the classes fence, traffic light, traffic sign, rider, truck, and motorcycle, which is possibly due to their characteristic depth profile learned during SDE. For example, a good depth estimation performance requires correctly segmenting poles or traffic signs as missing them can cause large depth errors. This can also be seen in Fig.~\ref{fig:examples}. DepthMix (XD) further improves the performance of wall, truck, bus, and train. This might be caused by the fact the DepthMix presents those rather difficult objects in another context, which might help the network to generalize better. 

In the suppl. materials, we further show that our method is still applicable if SDE is trained on a different dataset than semantic segmentation within a similar visual domain.

\subsection{Automatic Data Selection for Annotation}
\label{sec:results_label_selection}

\begin{table}
\caption{Comparison of data selection methods (DS: Diversity Sampling based on depth features, US: Uncertainty Sampling based on depth student error). mIoU in \%, std. dev. over 3 seeds.}
\label{tab:label_selection}
\vspace{0.2cm}
\centering
\begin{tabular}{llll}
\hline
\# Labeled          & 1/30 (100) & 1/8 (372) & 1/4 (744) \\
\hline\hline

Random & 48.75	\scriptsize{$\pm$1.61} & 59.14 \scriptsize{$\pm$1.02}	&	63.46	\scriptsize{$\pm$0.38}\\
Entropy & 53.63	\scriptsize{$\pm$0.77}	& 63.51	\scriptsize{$\pm$0.68}	& 66.18	\scriptsize{$\pm$0.50} \\
Ours (US) & 51.75	\scriptsize{$\pm$1.12}	& 62.77	\scriptsize{$\pm$0.46}	& 66.76	\scriptsize{$\pm$0.45} \\
Ours (DS) & 53.00	\scriptsize{$\pm$0.51}	& 63.23	\scriptsize{$\pm$0.69}	& 66.37	\scriptsize{$\pm$0.20} \\
Ours (DS+US) & \textbf{54.37}	\scriptsize{$\pm$0.36} & \textbf{64.25}	\scriptsize{$\pm$0.18}	& \textbf{66.94}	\scriptsize{$\pm$0.59} \\
\hline
\end{tabular}
\end{table}

Finally, we evaluate the proposed automatic data selection. Tab.~\ref{tab:label_selection} shows a comparison of our method with a baseline and a competing method. The baseline selects the labeled samples randomly, while the second, strong competitor uses active learning and iteratively chooses the samples with the highest segmentation entropy. In contrast to our method, this requires a human in the loop to create the semantic labels for iteratively selected images. It can be seen that our method with the combined Diversity Sampling and Uncertainty Sampling (DS+US) outperforms both comparison methods, demonstrating the effectiveness of ensuring diversity and exploiting difficult samples based on depth. It also supports the assumption that depth estimation and semantic segmentation are correlated in terms of sample difficulty. The class-wise analysis (see the last row of Fig.~\ref{fig:classwise_improvement}) shows that data selection significantly improves the performance of truck, bus, and train, which are usually difficult to distinguish in a semi-supervised setting. We would like to note that our automatic data selection method can be applied to any semantic segmentation method. 

\section{Conclusion}
\label{sec:conclusion}

In this work, we have studied how self-supervised depth estimation (SDE) can be utilized to improve semantic segmentation in both the semi-supervised and the fully-supervised setting. We introduced three effective strategies capable of leveraging the knowledge learned from SDE. First, we show that the SDE feature representation can be transferred to semantic segmentation, by means of SDE pretraining and joint learning of segmentation and depth. 
Second, we demonstrate that the proposed DepthMix strategy outperforms related mixing strategies by avoiding inconsistent geometry of the generated images.
Third, we present an automatic data selection for annotation algorithm based on SDE, which does not require human-in-the-loop annotations. 
We validate the benefits of the three components by extensive experiments on Cityscapes, where we demonstrate significant gains over the baselines and competing methods. By using SDE, our approach achieves state-of-the-art performance, suggesting that SDE can be a valuable self-supervision for semantic segmentation.

\vspace{2mm}
\noindent
\textbf{Acknowledgements}: This work is funded by Toyota Motor Europe via the research project TRACE-Zurich and by a research project from armasuisse.
\FloatBarrier

{\small
\bibliographystyle{ieee_fullname}
\bibliography{egbib}
}

\clearpage

\renewcommand\thesection{\Alph{section}}
\renewcommand{\thefigure}{S\arabic{figure}}
\renewcommand{\thetable}{S\arabic{table}}
\setcounter{section}{0}

\section{Further Implementation Details}

In the following paragraphs, a more detailed description of the network architecture and the training is provided. The reference implementation is available at \url{https://github.com/lhoyer/improving_segmentation_with_selfsupervised_depth}.

\paragraph{Network Architecture}

The neural network combines a DeepLabv3~\cite{chen2017deeplab} with a U-Net~\cite{ronneberger2015u} decoder for depth and segmentation prediction each. As encoder, a ResNet101 with dilated (instead of strided) convolutions in the last block is used, following~\cite{chen2017deeplab}. Features from multiple scales are aggregated by an ASPP~\cite{chen2017deeplab} block with dilation rates of 6, 12, and 18.
Similar to U-Net~\cite{ronneberger2015u}, the decoder has five upsampling blocks with skip connections. Each upsampling block consists of a 3x3 convolution layer (except the first block, which is the ASPP), a bilinear upsampling operation, a concatenation with the encoder features of the corresponding size (skip connection), and another 3x3 convolution layer. Both convolutional layers are followed by an ELU non-linearity. The number of output channels for the blocks are 256, 256, 128, 128, and 64. The last four blocks also have another 3x3 convolutional layer followed by a sigmoid activation attached to their output for the purpose of predicting the disparity at the respective scale. For effective multi-task learning, we additionally follow PAD-Net~\cite{xu2018pad} and deploy an attention-guided multi-modal distillation module with additional side output for semantic segmentation after the third decoder block. In experiments without multi-task learning, only the semantic segmentation decoder is used.
For pose estimation, we use a lightweight ResNet18 encoder followed by four convolutions to produce the translation and the rotation in angle-axis representation as suggested in \cite{godard2019digging}.

\paragraph{Runtime}

To give an impression of the computational complexity of our architecture, we provide the training time per iteration and the inference time per image on an Nvidia Tesla P100 in Tab.~\ref{tab:runtime}. The values are averaged over 100 iterations or 500 images, respectively. Please note that these timings include the computational overhead of the training framework such as logging and validation metric calculation.

\begin{table}
\caption{Training and inference time on an Nvidia Tesla P100 averaged over 100 iterations or 500 images, respectively. D-T: SDE Transfer Learning, D-M SDE Transfer and Multi-Task Learning, P:~Pseudo-Labelling, X-D: Mix Depth}
\label{tab:runtime}
\vspace{0.2cm}
\centering
\setlength{\tabcolsep}{5pt}
\begin{tabular}{lllllll}
\hline
D & P & X & Training Time & Inference Time\\\hline\hline
T &   &   & 188 ms/it & 66 ms/img\\
T & \checkmark  &   & 466 ms/it & 67 ms/img\\
T & \checkmark  & D & 476 ms/it & 66 ms/img\\
M & \checkmark  & D & 1215 ms/it & 160 ms/img\\
\hline
\end{tabular}
\end{table}

\paragraph{Data Selection}

In the data selection experiment, we use a slimmed network architecture for $f_{SIDE}$ with a ResNet50 backbone, 256, 128, 128, 64, and 64 decoder channels, and BatchNorm~\cite{ioffe2015batch} in the decoder for efficiency and faster convergence. The depth student network is trained using a berHu loss~\cite{zwald2012berhu, laina2016deeper}. The quality of the selected subset with annotations $\mathcal{G}_A$ is evaluated for semantic segmentation using our default architecture and training hyperparameters.

\section{Cross-Dataset Transfer Learning}

\begin{table*}
\caption{Performance on the CamVid test set (mIoU in \%, standard deviation over 3 random seeds). The SDE is trained on Cityscapes sequences. DT: SDE Transfer Learning, XD - DepthMix, S: Data Selection.}
\label{tab:camvid}
\vspace{0.2cm}
\centering
\setlength{\tabcolsep}{5pt}
\begin{tabular}{lllllll}
\hline
\# Labeled & 50    &        & 100     &       & 367 (Full) &       \\\hline\hline
Baseline & 59.16	\scriptsize{$\pm$1.79} & \blarrow &		63.05	\scriptsize{$\pm$0.59} & \blarrow &		68.18	\scriptsize{$\pm$0.13} & \blarrow \\
Ours (DT) & 62.75	\scriptsize{$\pm$2.32} &	+3.60 &	66.19	\scriptsize{$\pm$0.96} &	+3.15 &	70.45	\scriptsize{$\pm$0.35} &	+2.27 \\
ClassMix~\cite{olsson2020classmix} & 65.89	\scriptsize{$\pm$0.33} &	+6.73 &	67.48	\scriptsize{$\pm$1.02} &	+4.43 &	-	&  \\
Ours (DT+XD) & 66.82	\scriptsize{$\pm$1.16} &	+7.66 &	68.91	\scriptsize{$\pm$0.62} &	+5.86 &	71.46	\scriptsize{$\pm$0.22} &	+3.29 \\
Ours (DT+XD+S) & 68.23	\scriptsize{$\pm$0.39} &	+9.07 &	69.62	\scriptsize{$\pm$0.64} &	+6.57 &	- & \\
\hline
\end{tabular}
\end{table*}

In this section, we show that the unlabeled image sequences and the labeled segmentations can also originate from different datasets within similar visual domains. For that purpose, we train the SDE on Cityscapes sequences and learn the semi-supervised semantic segmentation on the CamVid dataset~\cite{brostow2009semantic}, which contains 367 train, 101 validation, and 233 test images with dense semantic segmentation labels for 11 classes from street scenes in Cambridge. To ensure a similar feature resolution, we upsample the CamVid images from $480 \times 360$ to $672 \times 512$ pixels and randomly crop to a size of $512 \times 512$.

Table~\ref{tab:camvid} shows that the results on CamVid are similar to our main results on Cityscapes. For $50$ labeled training samples, SDE pretraining improves the mIoU by $3.6$ percentage points, pseudo-labels and DepthMix by another $4.07$ percentage points, and data selection by another $1.41$ percentage points. In the end, our proposed method significantly outperforms ClassMix by $2.34$ percentage points for $50$ labeled samples and $2.14$ percentage points for $100$ labeled samples. Also for the fully labeled dataset, our method can improve the performance by $3.29$ percentage points.

\section{Further Example Predictions}
\label{sec:further_examples}

\begin{figure*}
    \centering
    \includegraphics[width=\linewidth]{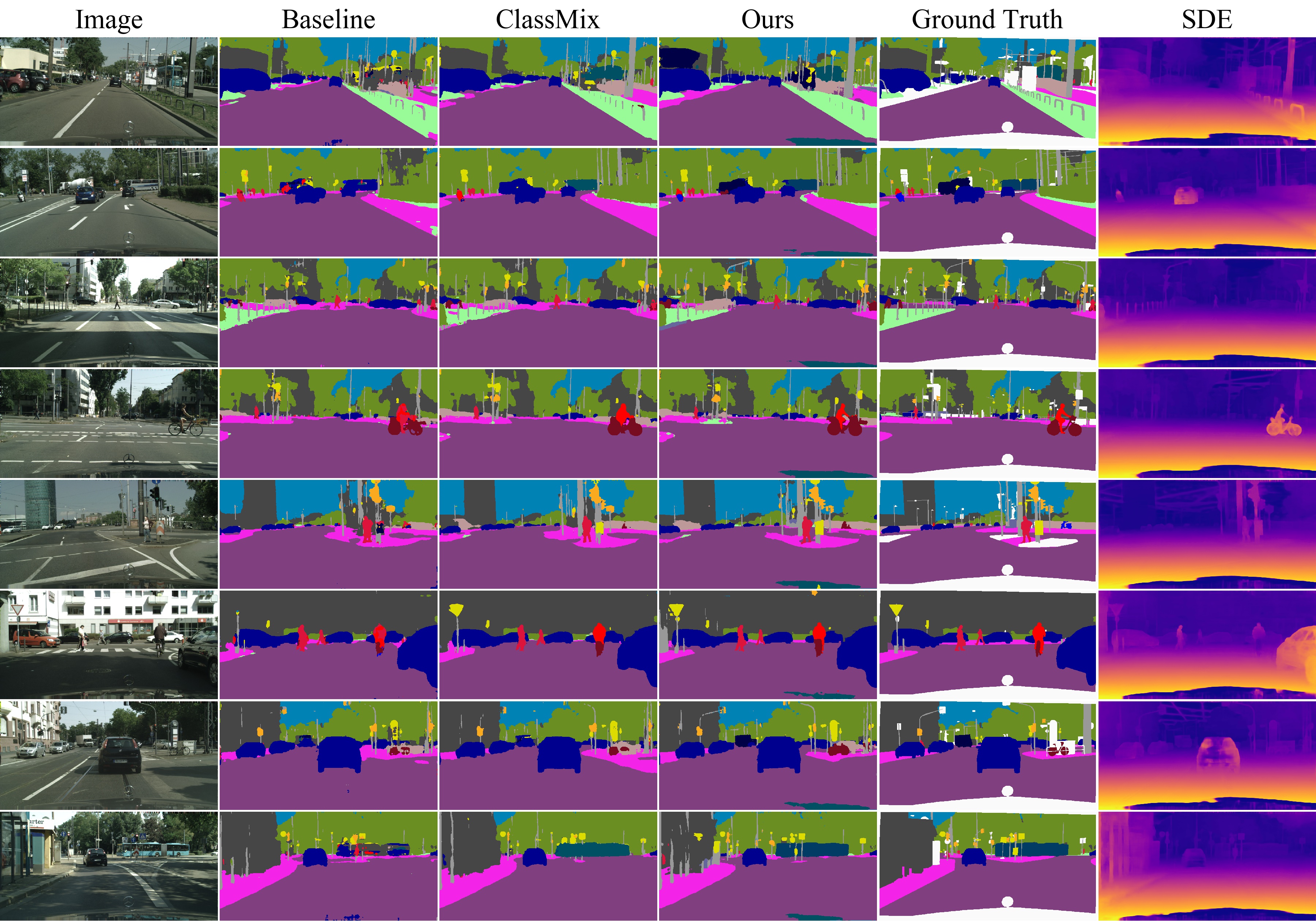}
    \caption{Further example predictions for 100 annotated training samples including the self-supervised disparity estimate of the multi-task learning framework.}
    \label{fig:further_examples}
\end{figure*}

Further examples for semantic segmentation and SDE are shown in Fig.~\ref{fig:further_examples}. In general, the same observations as in the main paper can be made. Our method provides clearer segmentation contours for objects that are bordered by pronounced depth discontinuities such as pole, traffic sign, or traffic light. We also show improved differentiation between similar classes such as truck, bus, and train. On the downside, SDE sometimes fails for cars driving directly in front of the camera (see 7th row in Fig.~\ref{fig:further_examples}) and violating the reconstruction assumptions. Those cars are observed at the exact same location across the image sequence and can not be correctly reconstructed during SDE training, even with correct depth and pose estimates. However, this differentiation between moving and non-moving cars does not hinder the transfer of SDE-learned features to semantic segmentation but can cause problems with DepthMix (see Section~\ref{sec:depthmix_examples}).

\section{DepthMix Real-World Examples}
\label{sec:depthmix_examples}

\begin{figure*}
    \centering
    a) \includegraphics[width=0.96\linewidth,valign=c]{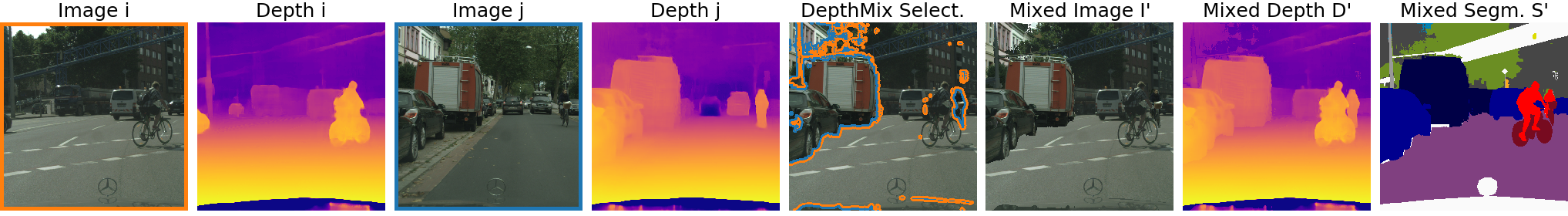}\\
    b) \includegraphics[width=0.96\linewidth,valign=c]{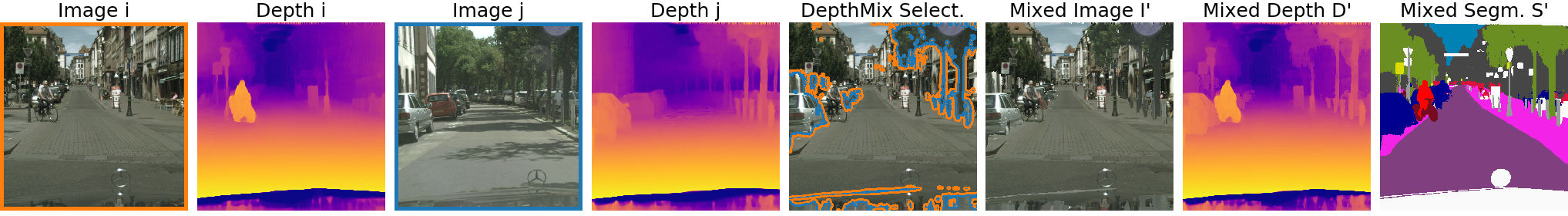}\\
    c) \includegraphics[width=0.96\linewidth,valign=c]{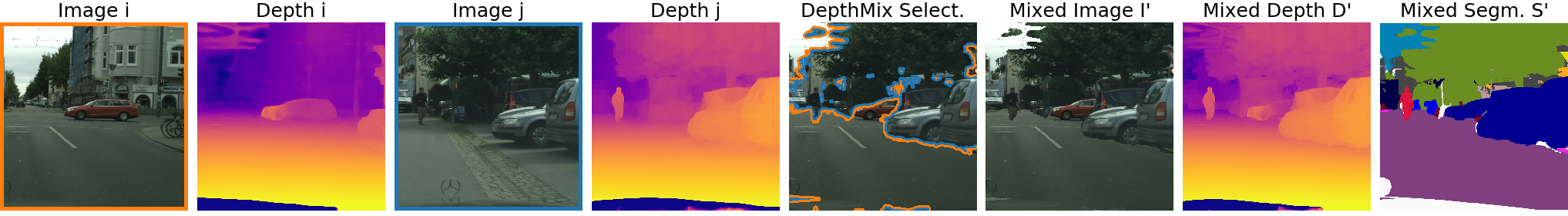}\\
    d) \includegraphics[width=0.96\linewidth,valign=c]{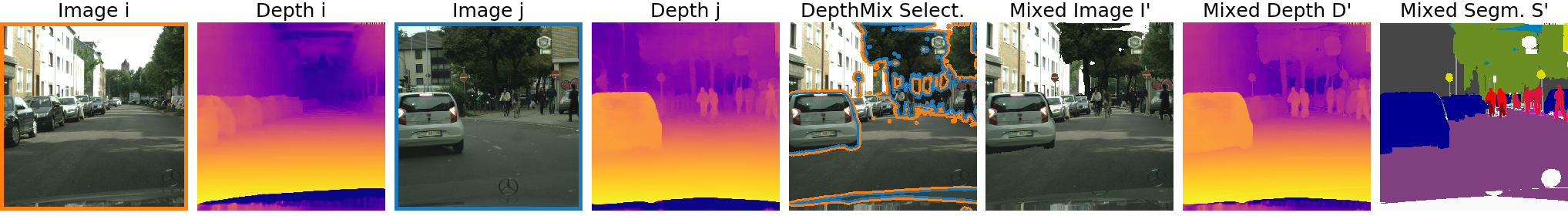}\\
    e) \includegraphics[width=0.96\linewidth,valign=c]{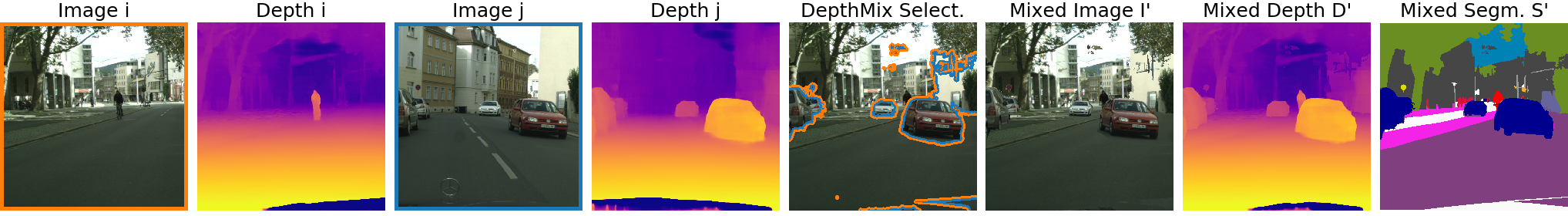}\\
    f) \includegraphics[width=0.96\linewidth,valign=c]{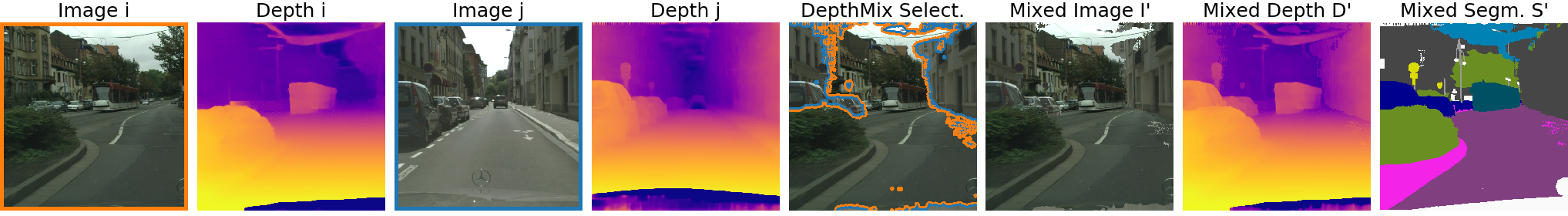}\\
    g) \includegraphics[width=0.96\linewidth,valign=c]{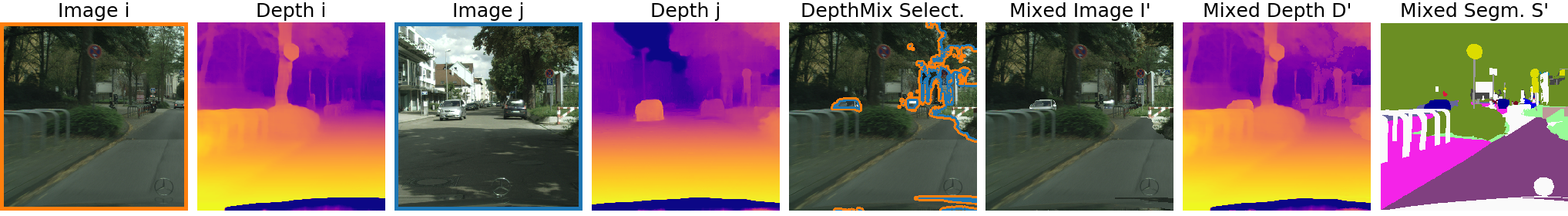}\\
    h) \includegraphics[width=0.96\linewidth,valign=c]{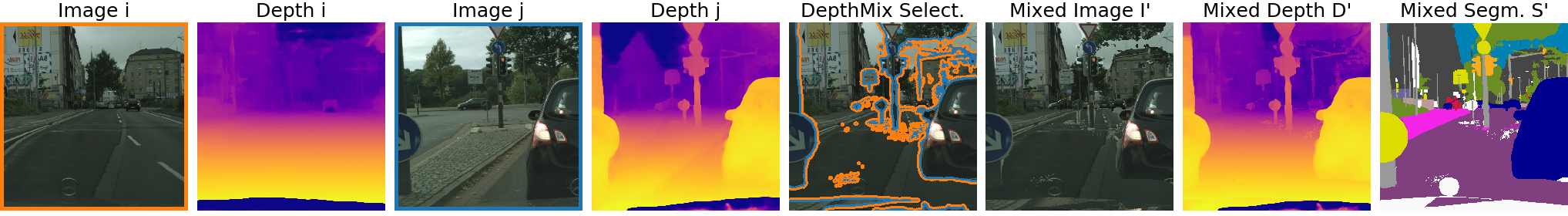}\\
    \caption{DepthMix applied to Cityscapes crops. From left to right, the source images with their SDE estimate, the mixed image $I'$ overlaid with border of the mix mask $M$ in blue/orange depending on the adjacent source image (i - orange, j - blue), the mixed image without visual guidance $I'$, the mixed depth $D'$, and the mixed segmentation $S'$ are shown. For simplicity, the source segmentations for the mixed segmentation $S'$ originate from the ground truth labels.}
    \label{fig:depthmix_examples}
\end{figure*}

In Fig.~\ref{fig:depthmix_examples}, we show examples of DepthMix applied to Cityscapes crops. Generally, it can be seen that DepthMix works well in most cases. The self-supervised depth estimates allow to correctly model occlusions and the produced synthetic samples have a realistic appearance.

\begin{figure*}
    \centering
    a) \includegraphics[width=0.96\linewidth,valign=c]{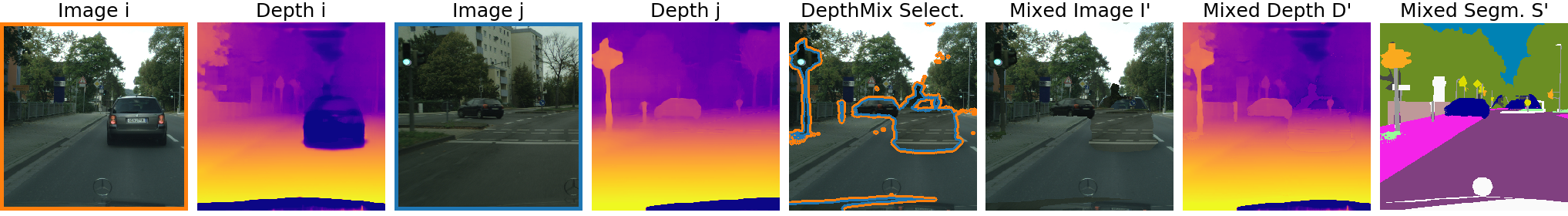}
    b) \includegraphics[width=0.96\linewidth,valign=c]{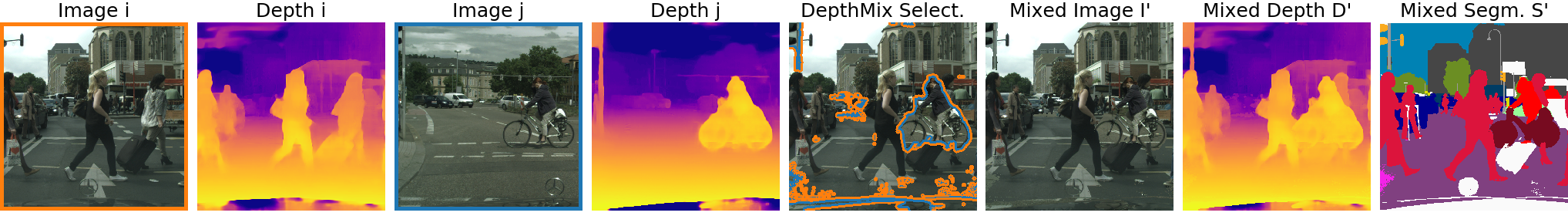}
    c) \includegraphics[width=0.96\linewidth,valign=c]{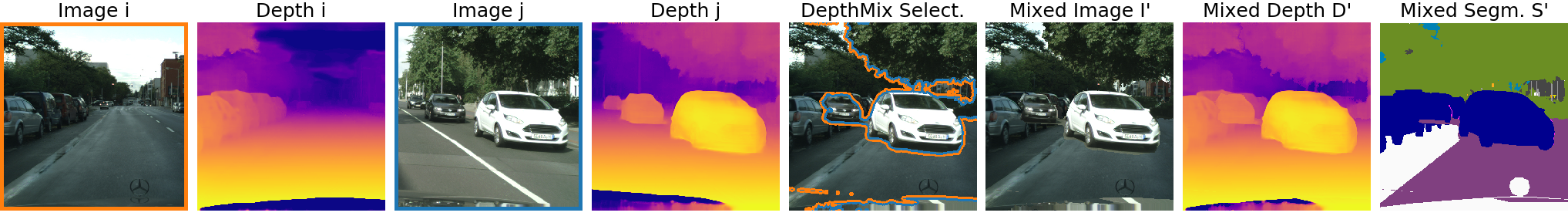}
    d) \includegraphics[width=0.96\linewidth,valign=c]{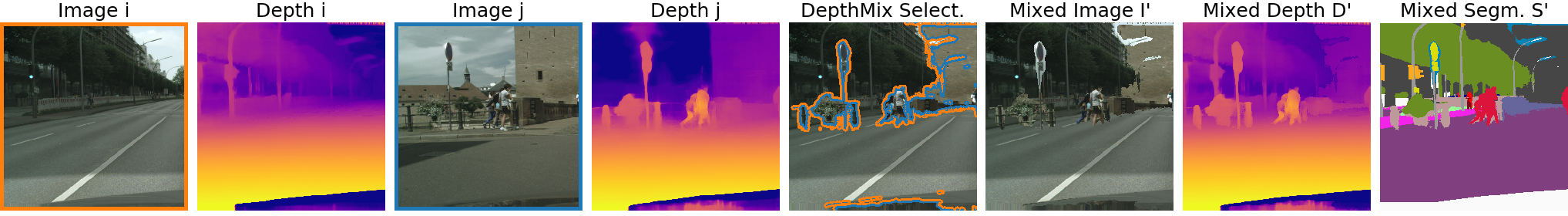}
    e) \includegraphics[width=0.96\linewidth,valign=c]{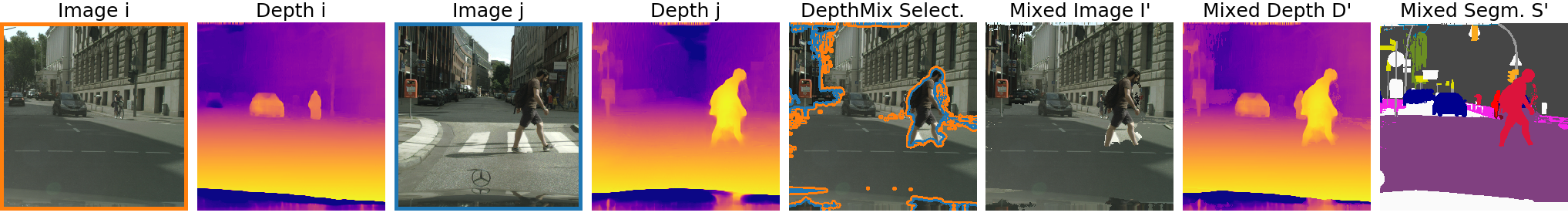}
    \caption{DepthMix failure cases. From left to right, the source images with their SDE estimate, the mixed image $I'$ overlaid with border of the mix mask $M$ in blue/orange depending on the adjacent source image (i - orange, j - blue), the mixed image without visual guidance $I'$, the mixed depth $D'$, and the mixed segmentation $S'$ are shown. For simplicity, the source segmentations for the mixed segmentation $S'$ originate from the ground truth labels.}
    \label{fig:depthmix_failure_cases}
\end{figure*}

In Fig.~\ref{fig:depthmix_failure_cases}, we show a selection of typical failure cases of DepthMix. First, the SDE can be inaccurate for dynamic objects (see Sec.~\ref{sec:further_examples}), which can cause an inaccurate structure within the mixed image (Fig.~\ref{fig:depthmix_failure_cases} a, b, and c). However, this type of failure case is common in ClassMix and its frequency is greatly reduced with DepthMix. A remedy might be SDE extensions that incorporate the motion of dynamic objects~\cite{casser2019depth, dai2020self, klingner2020self}.
Second, in some cases, the SDE can be imprecise and the depth discontinuities do not appear at the same location as the class border. This can cause artifacts in the mixed image (Fig.~\ref{fig:depthmix_failure_cases} d and e) but also in the mixed segmentation (Fig.~\ref{fig:depthmix_failure_cases} e: sky within the building). Note that the same can happen for ClassMix when using pseudo-labels for creating the mix mask.

\end{document}